\documentclass{article}
\usepackage[preprint]{neurips_data_2024}

\usepackage[utf8]{inputenc} 
\usepackage[T1]{fontenc}    
\usepackage{hyperref}       
\usepackage{url}            
\usepackage{booktabs}       
\usepackage{amsfonts}       
\usepackage{nicefrac}       
\usepackage{microtype}      
\usepackage{xcolor}         

\usepackage{graphicx}
\usepackage{amsmath}
\usepackage{amsfonts} 
\usepackage{tcolorbox}
\usepackage{tikz}  
\usepackage{paralist}
\usepackage{array}
\usepackage{makecell}
\usepackage{pifont} 
\usepackage{algorithm}
\usepackage{algorithmic}
\usepackage{subfigure}

\newcommand{\cmark}{\ding{51}}  
\newcommand{\xmark}{\ding{55}}  

\title{{\vspace{-0.9em}\includegraphics[width=0.3\textwidth]{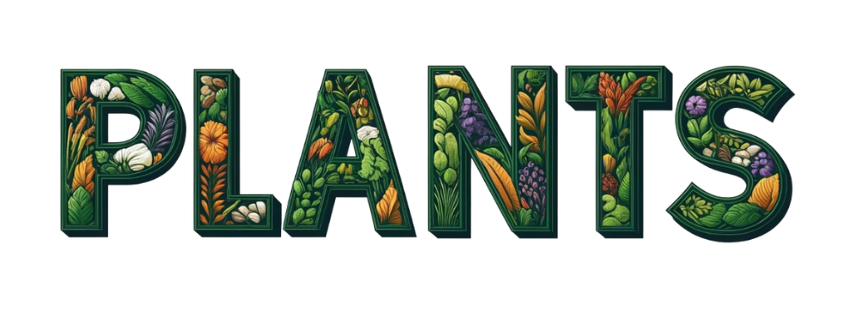}}\\ \vspace{-0.3em}A Novel Problem and Dataset for Summarization of \textit{Planning-Like} (PL) Tasks}

\author{%
  Vishal Pallagani \\
  \small{University of South Carolina} \\
  \texttt{vishalp@mailbox.sc.edu} \\
  \And
  Biplav Srivastava \\
  \small{University of South Carolina} \\
  \texttt{biplav.s@sc.edu} \\
    \And
  Nitin Gupta \\
  \small{University of South Carolina} \\
  \texttt{niting@email.sc.edu} \\
}

\begin{document}

\maketitle

\begin{abstract}
Text summarization is a well-studied problem that deals with deriving insights from {\em{unstructured text}} consumed by humans, and it has found extensive business applications. However, many real-life tasks involve generating a series of actions to achieve specific goals, such as workflows, recipes, dialogs, and travel plans. We refer to them as {\em planning-like} (PL)  tasks noting that the main commonality  they share is  control flow information. which may be partially specified. Their structure presents an opportunity to create more practical summaries to help users make quick decisions. We investigate this observation by introducing a novel plan summarization problem, presenting a dataset, and providing a baseline method for generating PL summaries. Using quantitative metrics and qualitative user studies to establish baselines, we evaluate the plan summaries from our method and large language models. We believe the novel problem and dataset can reinvigorate research in summarization, which some consider as a solved problem.
\end{abstract}

\section{Introduction}

Text summarization is a crucial task in natural language processing (NLP) that focuses on condensing large volumes of unstructured text into concise and informative summaries \citep{luhn1958automatic}. This task has significant applications in various domains such as news aggregation, document summarization, and content recommendation systems \citep{el2021automatic}. Traditional summarization techniques can be broadly categorized into extractive \citep{gupta2010survey} and abstractive methods \citep{gupta2019abstractive}. Extractive summarization selects key sentences or phrases from the original text, whereas abstractive summarization generates new sentences that capture the essence of the text. Recently, large language models (LLMs) have demonstrated remarkable capabilities, outperforming human summaries \citep{pu2023summarization} on several datasets such as Multi-News \citep{fabbri2019multi} and MediaSum \citep{zhu2021mediasum}.

Despite its extensive applications, text summarization has primarily concentrated on static documents, overlooking dynamic tasks that involve sequences of actions aimed at achieving specific goals. We refer to these tasks as \textit{planning-like} (PL) tasks\citep{srivastava2024case}. Examples of PL tasks include workflows, recipes, dialogs, and travel plans, which often contain control flow information critical for execution. For instance, consider the task of cooking a cheese sandwich. Numerous recipes exist for making a cheese sandwich, each with varying ingredients and steps. A summary for this PL task aims to condense these multiple recipes into a single, coherent summary. This summary would allow a knowledgeable user to quickly make a cheese sandwich based on the brief summary or help a user decide which recipe best suits their needs based on the ingredients they have available. This approach can be considered similar to multi-document summarization on a high level, where information from multiple sources is synthesized into a concise summary \citep{goldstein2000multi}. By summarizing multiple action sequences into coherent and actionable insights, we provide users with valuable information and facilitate quicker decision-making.

\begin{figure}[htbp]
    \centering
    \includegraphics[scale=0.63]{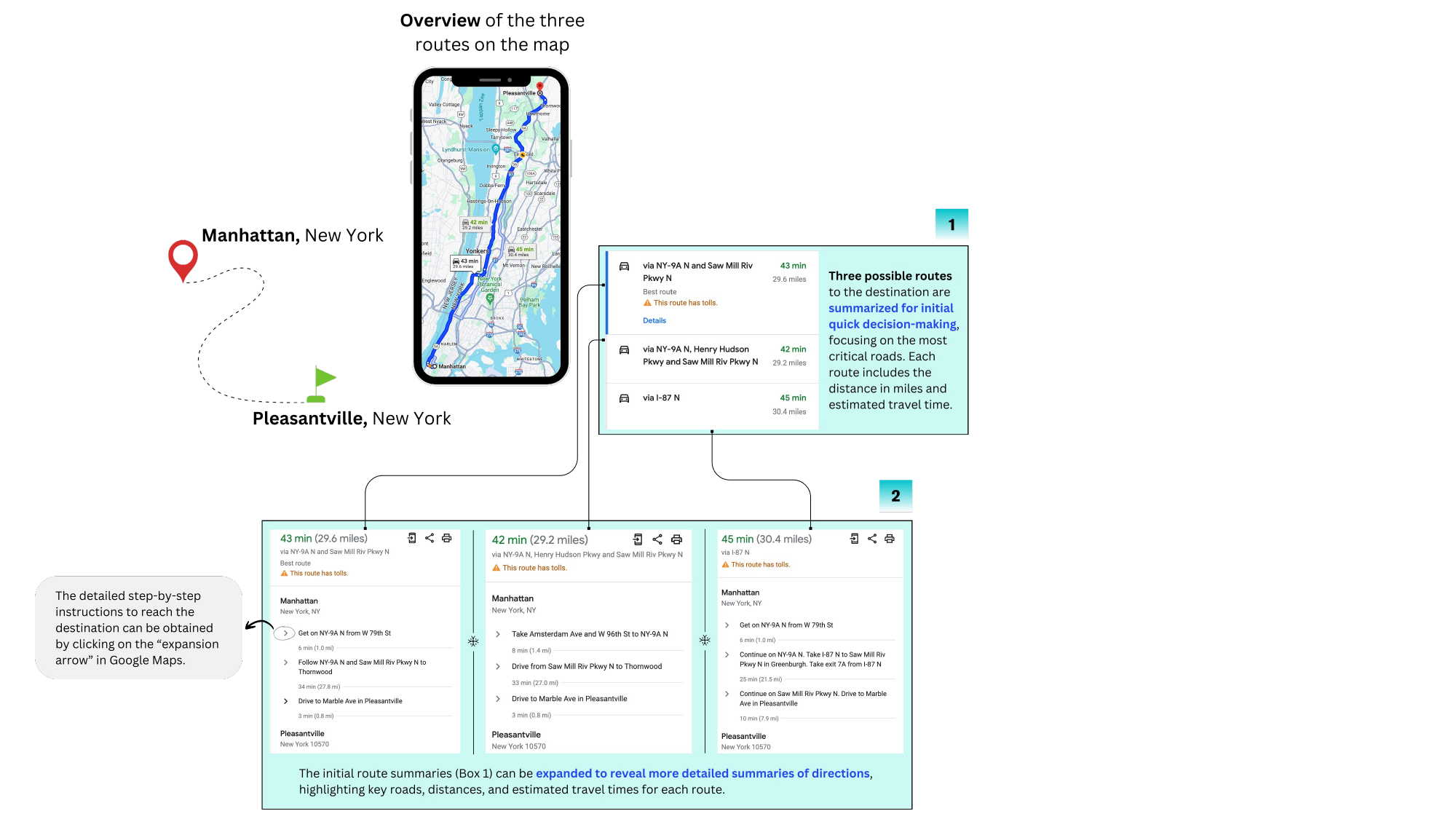}
    \caption{Google Maps summarizes three possible driving routes from Manhattan to Pleasantville, New York. The initial view (Box 1) includes key information like critical roads, estimated travel time, and distance, aiding quick decision-making. Detailed step-by-step directions can be accessed by expanding each summary present in Box 2.}
    \label{fig:googlemaps}
\end{figure}

Consider another example of routes from Google Maps, a commercial service offering travel routes between selected locations. In Figure \ref{fig:googlemaps}, we provide an instance where the user wants to find driving routes between Manhattan, New York, and Pleasantville, New York. Google Maps offers multiple route options visually on the map and provides a summary of three possible routes to reach the destination. This summary focuses on the critical roads, estimated travel time, and distance. This allows the user to choose their preferred route without going through the complete step-by-step instructions for all three options. Each summary in Box 1 can be expanded to reveal more detailed summaries, including additional key roads or waypoints. This capability enables quick decision-making and efficient route planning, illustrating the utility of summarization in PL tasks.

To address the gap in summarization literature for PL tasks, we introduce the novel problem of summarizing planning like (PL) tasks\footnote{We also refer to it as plan summarization or PL summaries.}. Plan summarization aims to create concise and coherent summaries of action sequences that achieve specific goals, thereby facilitating quick understanding and decision-making. Unlike traditional text summarization, plan summarization must account for the executability  and logical flow of actions.

We present a new dataset, called as PLANTS\footnote{https://github.com/VishalPallagani/PLANTS-benchmark}, specifically designed for plan summarization tasks, encompassing diverse domains such as automated plans, recipes, and travel plans. Additionally, we propose a baseline method for generating PL summaries. Our evaluation includes comparisons with summaries generated by both extractive and abstractive methods through a user study. We believe that introducing the plan summarization problem and providing a relevant dataset will spark renewed interest in the summarization research community. Our contributions are threefold: 
\begin{inparaenum}[(1)]
    \item Definition of the planning task summarization;
    \item Creation of a dataset tailored for PL tasks;
    \item Development of a baseline method for generating  summaries;
    \item Initial evaluation of how users perceive PL summaries from the baseline method and LLMs.
\end{inparaenum}

\section{\textit{Planning-like} Tasks}

\textit{Planning-like} tasks involve a series of actions required to achieve specific goals. These tasks are defined and explored in \citet{srivastava2024case}. In this paper, we focus on three primary domains of PL tasks: automated plans, recipes, and travel routes. Each of these domains involves unique challenges and characteristics that necessitate effective summarization for better user comprehension and decision-making.

\noindent \textbf{Automated Plans} \\
Automated planning \citep{ghallab2004automated} involves creating action sequences for intelligent agents to achieve specified goals. In automated planning, a problem is typically represented as a tuple consisting of states, actions, and goals. The objective is to generate an automated plan that transitions the system from the initial state to the goal state while satisfying certain constraints. The semantics of automated plans require them to be sound and feasible, meaning each action must be executable in the given context, and the sequence must logically lead to the achievement of the goal. Summarizing automated plans helps in quickly understanding the essential steps and ensuring all actions are executable.

\noindent \textbf{Recipes}\\
In the domain of culinary arts, recipes are structured sequences of actions aimed at preparing specific dishes. Each recipe includes a list of ingredients and step-by-step instructions for combining them. Given the multitude of recipes available for a single dish, there can be significant variation in ingredients and preparation methods. This diversity makes it challenging for users to quickly identify the essential components and steps needed to prepare a dish. Summarizing recipes allows users to identify must-have ingredients and critical steps, making it easier to choose or adapt a recipe based on available ingredients.

\noindent \textbf{Travel Routes}\\
Travel planning involves creating efficient paths from a starting location to a destination. This process includes determining the optimal route, considering factors such as distance, travel time, and road conditions. Travel routes are complex, often involving multiple possible paths and decisions about which roads or highways to take. Summarizing travel routes provides a clear overview of the main paths, travel times, and distances, aiding in quick decision-making and efficient route planning.

These PL tasks, as summarized in Table \ref{tab:pltasks}, highlight the different characteristics and requirements across domains. Summarizing these tasks enhances usability and accessibility, providing users with concise, actionable insights for efficient decision-making and task execution.

\begin{table}[h!]
\centering
\caption{Characterizing \textit{Planning-like} Tasks.}
\resizebox{\textwidth}{!}{
\begin{tabular}{@{}p{2cm} p{3.5cm} c c c c p{4.5cm}@{}}
\toprule
Domain & \makecell{State \\ Representation} & \makecell{Control \\ Flow} & \makecell{Data \\ Flow} & \makecell{Auto \\ Generation} & \makecell{Auto \\ Execution} & Comments\\ 
\midrule
Automated Plans & Full initial state, partial goal state & \centering \textcolor{green}{\cmark} & \centering Minimal & \centering \textcolor{green}{\cmark} & \centering \textcolor{green}{\cmark} & Precise action sequences ensuring sound execution.\\ 
\midrule
Recipes & List of ingredients and steps & \centering \textcolor{green}{\cmark} & \centering Moderate & \centering \textcolor{red}{\xmark} & \centering \textcolor{red}{\xmark} & Structured instructions for food preparation with variations across different recipes.\\ 
\midrule
Travel Routes & Start and destination points & \centering \textcolor{green}{\cmark} & \centering Extensive & \centering \textcolor{green}{\cmark} & \centering \textcolor{red}{\xmark} & Step-by-step travel paths with critical roads, travel times, and distances to reach the destination.\\ 
\bottomrule
\end{tabular}
}
\label{tab:pltasks}
\end{table}
\section{Planning Task Summarization}

Planning task summarization involves generating a concise summary of multiple plans that achieve the same goal. In various domains, such as travel planning, recipe generation, and automated planning, it is common to have multiple possible plans to reach a desired outcome. Each plan may differ in the sequence and number of actions required. Inspired by early work on process summarization \citep{srivastava2010summarizing}, our approach aims to enhance user comprehension and facilitate better decision-making by providing a summary that consolidates these multiple plans into a single, coherent overview, highlighting the key actions and considerations for achieving the goal.

We formally define the planning task summarization problem as follows. Given a set of plans \( P = \{ p_1, p_2, \ldots, p_n \} \), each plan \( p_i \) consists of a sequence of actions \( \{ a_{i1}, a_{i2}, \ldots, a_{im} \} \) designed to achieve a common goal \( G \). The task is to produce a summary plan \( P^* \) that is a function of the size and number of actions constrained by metadata. Mathematically, this can be expressed as:

\[
P^* = \text{Summarize}(P, \text{constraints})
\]

where the constraints may be in terms of textual features (e.g.,  maximum allowable characters, words or lines) or plan features (e.g., maximum number of actions) in the summary plan. Hence, it is expected that |$P^*$| << |$P$|. These constraints ensure that the summary plan remains concise and focused on the most critical actions necessary to achieve the goal.

Several challenges arise in the planning task summarization process. Different plans might take varied approaches to achieve the same goal, making it challenging to create a summary that captures the essential steps without losing critical diversity. Additionally, the summary must strictly adhere to the provided constraints, ensuring it remains concise and relevant. Another significant challenge is the selection of actions from the original plans to include in the summary. The goal is to ensure that the summary is representative of the original plans and efficient in the number of actions.

\section{PLANTS Dataset}
\begin{figure*}
    \centering
    \includegraphics[width=0.9\textwidth]{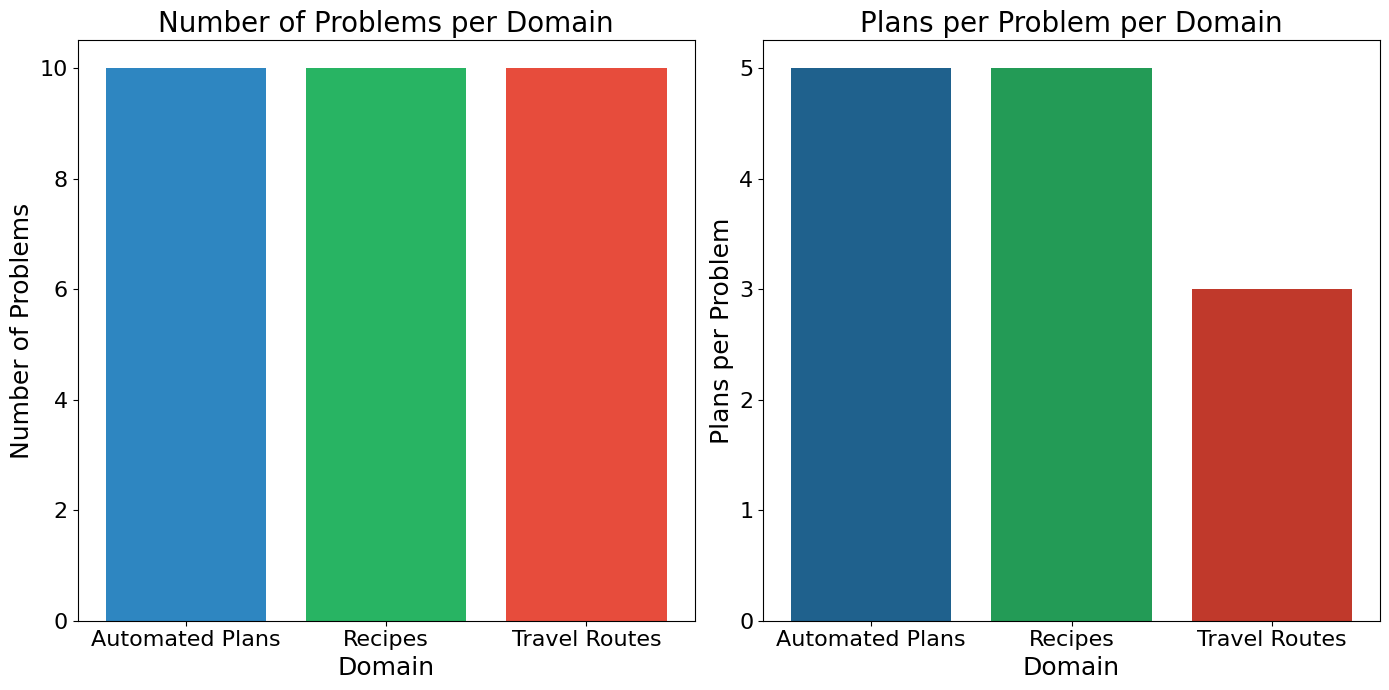}
    \caption{Distribution of problems and plans across domains. \textbf{Left:} shows the number of problems per domain, with each domain having 10 problems. \textbf{Right:} displays the average number of plans per problem for each domain.}
    \label{fig:plants-stats}
\end{figure*}

In this section, we introduce the PLANTS dataset, specifically designed for planning task summarization. The dataset encompasses three distinct planning-like tasks: automated plans, recipes, and travel routes. For each task, we have curated 10 different problems/goals. Each goal has 5 different plans for automated plans and recipes, and 3 different plans for travel routes, resulting in a total of 130 diverse plans in the dataset (see Figure \ref{fig:plants-stats}).

\noindent \textbf{Automated Plans:} For generating automated plans, we utilized five classical planning domains from the \texttt{downward-benchmarks}\citep{downward_benchmarks}: \texttt{blocks}, \texttt{driverlog}, \texttt{mprime}, \texttt{openstacks-strips}, and \texttt{queen-split}. These domains are released as part of the International Planning Competition (IPC) \citep{ipc}. The \texttt{downward-benchmarks} repository includes both the domains and their corresponding problems, where the goals are defined. We selected two distinct problems (i.e., goals) from each planning domain, resulting in a total of ten unique goals. Each problem was solved using SymK \citep{speck2020symbolic}, a state-of-the-art classical optimal and top-k planner based on symbolic search that extends Fast Downward \citep{helmert2006fast}. We set k to 5, generating five different plans for each problem. This approach ensures that our dataset contains a variety of viable solutions for each planning problem, providing a robust basis for summarization.

\noindent \textbf{Recipes:} For the recipes, we manually selected ten distinct and commonly made dishes such as cheese sandwich, guacamole and omelette from the Recipe1M+ dataset \citep{marin2021recipe1m+}. Recipe1M+ is a large-scale dataset containing over one million recipes with associated images and instructions. \underline{\textit{Assumption:}} To ensure diversity in preparation methods, we assume that distinct ingredient lists will result in different preparation steps. Based on this assumption, we extracted five different recipes for each dish by calculating the Jaccard similarity between the ingredient lists and selecting recipes with low similarity scores. This method ensures that the chosen recipes have varied ingredients, leading to diverse preparation steps. Specifically, we only extracted the ingredients and step-by-step instructions for each recipe. This manual selection and extraction process ensures that our dataset includes multiple viable approaches to achieve the same culinary goal, providing a robust basis for summarization.

\noindent \textbf{Travel Routes:} For the travel routes, we manually selected ten different pairs of start and destination coordinates to ensure a diverse set of route planning problems. The coordinates were chosen to cover a variety of urban layouts, providing a comprehensive testbed for summarization. We utilized the OpenStreetMap (OSM) API \citep{haklay2008openstreetmap}, a collaborative mapping project that provides free geographic data and mapping services, to generate routes between these coordinates. The OSM API allows for the extraction of detailed route information, including road networks and step-by-step directions. For each pair of coordinates, the API generates atmost three distinct routes, ensuring that the routes are unique by default. We extracted the step-by-step directions for each route, including the sequence of roads and waypoints. This approach ensures that our dataset captures a variety of viable travel options for each route planning problem.

\section{Experimental Settings}

In this section, we describe the different models used for plan summary generation and also discuss the user study settings. The constraints applied to these models and the prompt templates used for GPT-4o are detailed in Supplementary Material (Section 3).

\subsection{Models}

For each task, we use GPT-4o as the representative of LLMs and an abstractive technique for obtaining plan summaries. For extractive summarization, we use TextRank. Additionally, we developed a new frequency-based baseline method for extractive plan summarization. Each approach receives as input a set of plans to generate a summary. For automated plans and recipes, each set contains 5 plans, and for travel routes, each set contains 3 plans.

\begin{algorithm}
\caption{Baseline: Plan Summary Generation}
\label{alg:plan_summary}
\begin{algorithmic}[1]
\REQUIRE List of plans \textit{data}, each plan is a list of actions
\ENSURE Summary of the planning task problem

\STATE \textbf{Function} \textit{parse\_data(data)}
    \STATE Initialize \textit{parsed\_data} as an empty list
    \FOR{each plan in \textit{data}}
        \STATE Parse actions in the plan and add to \textit{parsed\_data}
    \ENDFOR
    \RETURN \textit{parsed\_data}

\STATE \textbf{Function} \textit{ngrams(lst, n)}
    \STATE Generate and return n-grams from list \textit{lst}

\STATE \textbf{Function} \textit{analyze\_text\_view(parsed\_data, ngram\_size)}
    \STATE Initialize \textit{all\_items} as an empty list
    \FOR{each action in \textit{parsed\_data}}
        \STATE Add action to \textit{all\_items}
    \ENDFOR
    \STATE Count and filter items and n-grams in \textit{all\_items}
    \RETURN Filtered items and n-grams

\STATE \textbf{Function} \textit{analyze\_plan\_view(parsed\_data)}
    \STATE Extract and count actions and secondary mentions in \textit{parsed\_data}
    \STATE Find the shortest plan and most common action sequence
    \RETURN Analysis of plan view

\STATE \textbf{Function} \textit{generate\_summary(text\_view, item\_view)}
    \STATE Summarize common actions, secondary mentions, shortest plan, and common action sequences
    \RETURN Summary
\end{algorithmic}
\end{algorithm}

Algorithm \ref{alg:plan_summary} outlines our baseline method, which involves parsing the plans to extract actions and creating a structured representation of the data. This structured data is then analyzed in two views: text view and plan view. The text view analysis identifies common items and n-grams by counting the frequency of individual actions and sequences of actions. The plan view analysis examines the structure and sequence of actions, identifying the most common actions, secondary mentions (such as objects or ingredients), the shortest plan, and the most common action sequences. The results from these analyses are combined to generate a plan summary.

\subsection{User Study}
To assess the ease of understanding, clarity for action, and overall preference for the summaries, we conducted a human evaluation involving ten annotators. The annotators were students (undergraduate and graduate students) and faculty staff, all with an understanding of the three PL tasks: automated plans, recipes, and travel routes. For each PL task, we provided the annotators with the actual plans and presented them with summaries generated by three different methods: GPT-4 (abstractive), TextRank (extractive), and our frequency-based baseline method (extractive). To ensure the reliability of our results, we calculated the overall inter-annotator agreement using Cohen’s kappa coefficient \citep{cohen1968weighted}. We found that the agreement among annotators was acceptable, with a coefficient of 0.72.

\section{Experimental Results}

\subsection*{Experiment 1: Comparing the number of tokens across the summaries}
\begin{figure}
    \centering
    \includegraphics[width=0.7\textwidth]{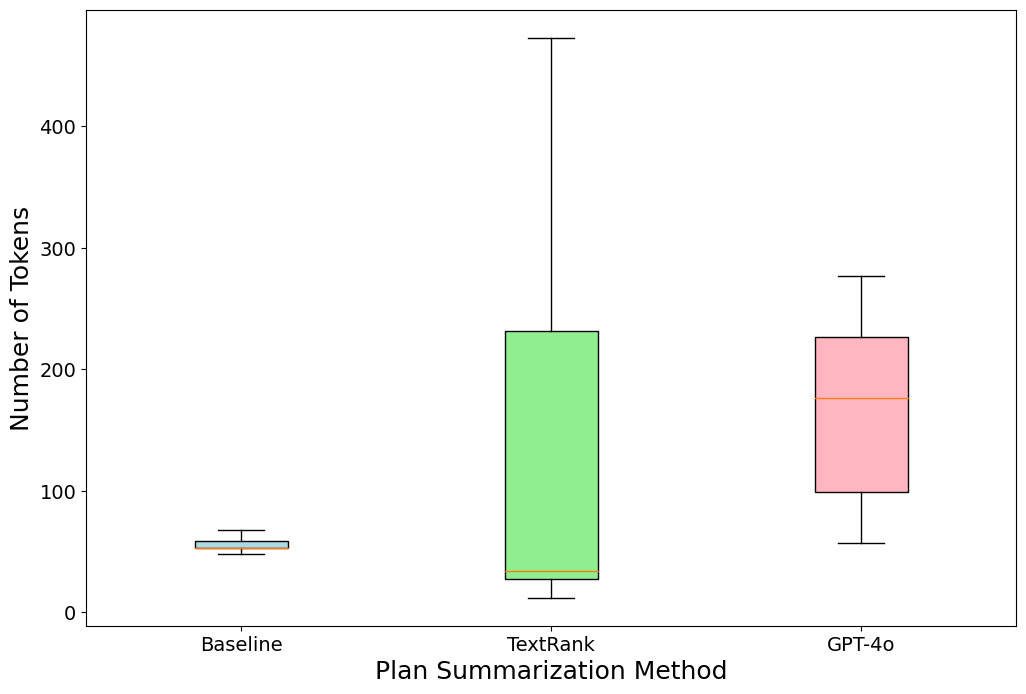}
    \caption{Comparison of token counts across different summarization approaches.}
    \label{fig:token-counts}
\end{figure}

Figure \ref{fig:token-counts} shows the boxplot comparing the token counts across three summarization methods: baseline, TextRank, and GPT-4o. The median token count for baseline is around 53, indicating consistent summary lengths with minimal variability. TextRank exhibits significant variability, with a median token count lower than baseline, reflecting diverse summary lengths. GPT-4o displays the highest median token count at approximately 176.5, indicating longer and more detailed summaries, with a wider interquartile range. This analysis highlights the differences in summary lengths, providing insights into the summarization characteristics of each method.

\subsection*{Experiment 2: Comparing the information-richness of the summaries}
\begin{figure}
    \centering
    \includegraphics[width=0.7\textwidth]{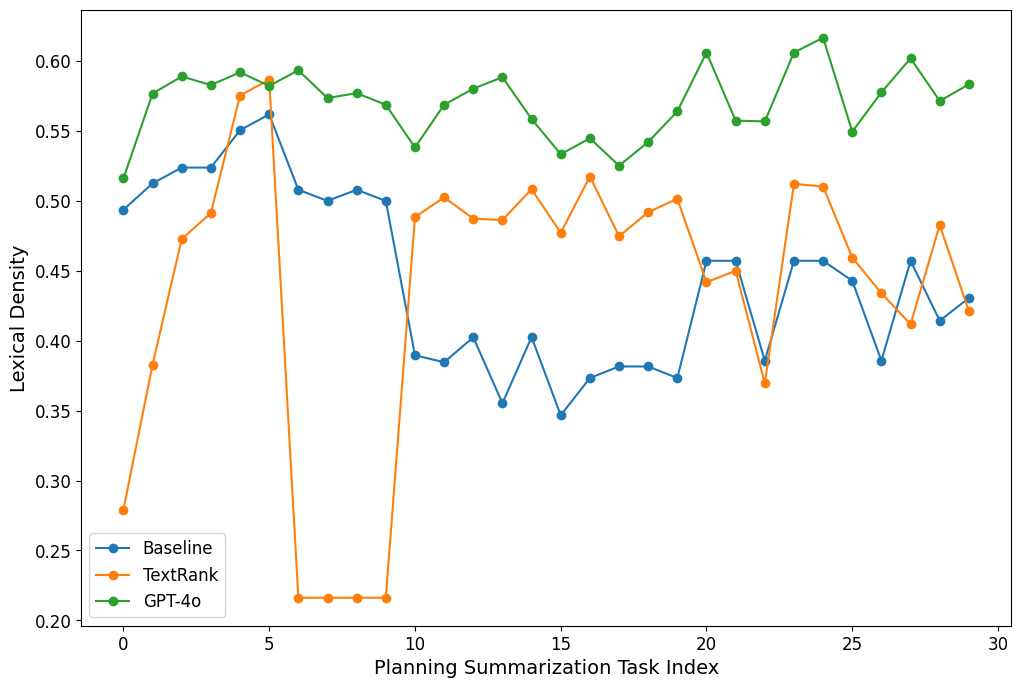}
    \caption{Comparison of lexical diversity across different summarization approaches to understand their information-richness.}
    \label{fig:lexical-div}
\end{figure}

In this experiment, we measure the lexical density of summaries generated by baseline, TextRank, and GPT-4o to evaluate their information richness. Lexical density is calculated as the proportion of content words—nouns, verbs, adjectives, and adverbs—to the total number of words in a summary. Figure \ref{fig:lexical-div} shows the lexical density of the three summary methods across 30 planning summarization tasks in the benchmark dataset. GPT-4o consistently achieves the highest lexical density, indicating it produces the most information-rich summaries. The baseline demonstrates moderate lexical density, followed by TextRank, which exhibits the lowest and most variable lexical density.

\subsection*{Experiment 3: Comparing the ease of understanding of the summaries}
From the user studies, we obtained results on how easy it is to understand a summary to take an action. Each summary was rated on a scale from 1 to 5, with 1 being very difficult to understand and 5 being very easy to understand. The average ease of understanding scores are presented in Table \ref{tab:ease_of_understanding}. GPT-4o received the highest ease of understanding scores across the three PL tasks. For automated plans, the baseline approach ranked second, while TextRank was rated second for recipes and travel routes.

\begin{table}[h]
    \centering
    \caption{Ease of understanding scores for the summaries across three different planning tasks.}
    \begin{tabular}{lccc}
        \toprule
        & \textbf{Baseline} & \textbf{TextRank} & \textbf{GPT-4o} \\
        \midrule
        \textbf{Automated Plans}   & 3.16 & 2.39 & \textbf{4.09} \\
        \textbf{Recipes}      & 2.77 & 3.41 & \textbf{4.68} \\
        \textbf{Travel Routes}& 2.70 & 3.45 & \textbf{3.99} \\
        \bottomrule
    \end{tabular}
    \label{tab:ease_of_understanding}
\end{table}

\subsection*{Experiment 4: User preference of the summaries}
The user study was also used to rank the summaries based on preferences. The aggregate preferences for each summary choice were then analyzed. For automated plans, GPT-4o was the first preference for 76\% of users, followed by the baseline approach as the second preference for 44\%, and TextRank as the third preference for 59\%, as shown in Table \ref{tab:order_preference_pl_tasks}. GPT-4o received the first preference across all three planning tasks, with TextRank and the baseline approach varying in their ranking depending on the specific task.

\begin{table}[h]
    \centering
    \caption{Order preference percentages for each summary across different PL tasks.}
    \begin{tabular}{lcccc}
        \toprule
        \textbf{PL Task} & \textbf{Summary} & \textbf{1st Preference} & \textbf{2nd Preference} & \textbf{3rd Preference} \\
        \midrule
        \textbf{Automated Plans} & Baseline & 15\% & \textbf{44\%} & 41\% \\
        & TextRank & 9\% & 32\% & \textbf{59\%} \\
        & GPT-4o   & \textbf{76\%} & 24\% & 0\% \\
        \midrule
        \textbf{Recipes} & Baseline & 10\% & 20\% & \textbf{70\%} \\
        & TextRank & 7\% & \textbf{67\%} & 26\% \\
        & GPT-4o & \textbf{83\%} & 13\% & 4\% \\
        \midrule
        \textbf{Travel Routes} & Baseline & 15\% & 13\% & \textbf{72\%} \\
        & TextRank & 34\% & \textbf{46\%} & 20\% \\
        & GPT-4o & \textbf{51\%} & 41\% & 8\% \\
        \bottomrule
    \end{tabular}
    \label{tab:order_preference_pl_tasks}
\end{table}
\section{Conclusion}
In this work, we introduced the novel problem of planning task summarization. To address this problem, we developed the PLANTS dataset, encompassing three distinct PL tasks: automated plans, recipes, and travel routes. Alongside the dataset, we also presented a frequency-based baseline method for plan summarization. We evaluated both abstractive and extractive summarization methods for planning task summarization through user studies and empirical analysis. Our findings indicate that while GPT-4o is the preferred approach for generating plan summaries due to its detailed and information-rich outputs, further evaluation is needed to verify if these summaries maintain the executional semantics of PL tasks. The issue of hallucination in abstractive methods remains a significant challenge that warrants further investigation. Additionally, there is a need to develop evaluation metrics specifically tailored for PL task summaries to ensure their effectiveness and reliability. 

We believe this work represents an initial effort towards advancing research in planning task summarization. The broader impact of this research could influence various domains, including robotics, dialog agents, and planning agents. We hope our contributions will inspire further advancements and exploration in this field, ultimately leading to more robust and efficient summarization techniques, datasets, and evaluation metrics for the problem of planning task summarization.

\section{Limitations}

\noindent \textbf{Size of the Dataset:} While the PLANTS dataset provides a valuable starting point for planning task summarization, it includes only 10 problems per domain, with 5 plans each for automated plans and recipes, and 3 plans each for travel routes. This limited size may not fully capture the variability and complexity of real-world planning tasks. Additionally, the dataset does not include gold summaries, as it is challenging to obtain authoritative summaries for PL tasks due to their inherent variability and subjective nature. However, to facilitate future research, we release the generators used to create this dataset, allowing for the development of larger and more diverse datasets across these domains.

\noindent \textbf{Evaluation Metrics:} The evaluation metrics employed in this study, such as human preference and ease of understanding, are inherently subjective and may not fully reflect the executional semantics of the plans.

\noindent \textbf{Inter-Annotator Agreement:} Although we measured inter-annotator agreement using Cohen’s kappa and found it to be acceptable, the subjective nature of human evaluation introduces potential variability in judgments. Future work could explore more rigorous training for annotators.

\section{Ethics Statement}
The development and evaluation of the PLANTS dataset were conducted with strict adherence to ethical standards. All data were sourced from publicly available repositories, ensuring compliance with usage terms and privacy regulations. Human evaluators, consisting of graduate students and professors with domain expertise, participated voluntarily and provided informed consent. Their responses were anonymized to maintain privacy. The dataset and evaluation methods were designed to minimize bias and ensure accuracy. We release the dataset generators for research purposes, encouraging responsible use in compliance with ethical guidelines. This work aims to benefit multiple domains, including robotics and planning agents, and we advocate for the responsible deployment of summarization technologies to avoid potential harm.

\section*{Acknowledgements}
We would like to thank Amitava Das for discussions related to textual summarization and for helping us build parallels to planning task summarization.

\bibliographystyle{plainnat}
\bibliography{references.bib}

\section*{Checklist}

\begin{enumerate}

\item For all authors...
\begin{enumerate}
  \item Do the main claims made in the abstract and introduction accurately reflect the paper's contributions and scope?
    \answerYes{}
  \item Did you describe the limitations of your work?
    \answerYes{}
  \item Did you discuss any potential negative societal impacts of your work?
    \answerYes{}
  \item Have you read the ethics review guidelines and ensured that your paper conforms to them?
    \answerYes{}
\end{enumerate}

\item If you are including theoretical results...
\begin{enumerate}
  \item Did you state the full set of assumptions of all theoretical results?
    \answerNA{}
	\item Did you include complete proofs of all theoretical results?
    \answerNA{}
\end{enumerate}

\item If you ran experiments (e.g. for benchmarks)...
\begin{enumerate}
  \item Did you include the code, data, and instructions needed to reproduce the main experimental results (either in the supplemental material or as a URL)?
    \answerYes{}
  \item Did you specify all the training details (e.g., data splits, hyperparameters, how they were chosen)?
    \answerNA{}
	\item Did you report error bars (e.g., with respect to the random seed after running experiments multiple times)?
    \answerNA{}
	\item Did you include the total amount of compute and the type of resources used (e.g., type of GPUs, internal cluster, or cloud provider)?
    \answerYes{}
\end{enumerate}

\item If you are using existing assets (e.g., code, data, models) or curating/releasing new assets...
\begin{enumerate}
  \item If your work uses existing assets, did you cite the creators?
    \answerYes{}
  \item Did you mention the license of the assets?
    \answerYes{}
  \item Did you include any new assets either in the supplemental material or as a URL?
    \answerYes{}
  \item Did you discuss whether and how consent was obtained from people whose data you're using/curating?
    \answerYes{}
  \item Did you discuss whether the data you are using/curating contains personally identifiable information or offensive content?
    \answerYes{}
\end{enumerate}

\item If you used crowdsourcing or conducted research with human subjects...
\begin{enumerate}
  \item Did you include the full text of instructions given to participants and screenshots, if applicable?
    \answerYes{}
  \item Did you describe any potential participant risks, with links to Institutional Review Board (IRB) approvals, if applicable?
    \answerYes{}
  \item Did you include the estimated hourly wage paid to participants and the total amount spent on participant compensation?
    \answerNA{}
\end{enumerate}

\end{enumerate}


\end{document}


\section*{Supplementary Material}
\tableofcontents
\section{Essential Information}

\begin{center}
\begin{tikzpicture}
    \definecolor{bgcolor}{HTML}{F9F9F9}
    \definecolor{linkcolor}{HTML}{007ACC}
    \definecolor{boxcolor}{HTML}{EDEDED}

    \node[fill=boxcolor, text=linkcolor, align=left, rounded corners, draw=none, inner sep=8pt, font=\small, text width=6cm] at (4, 7.5) {
        \faGithub \ \href{https://github.com/VishalPallagani/PLANTS-benchmark}{\textbf{GitHub Repository}} \\
        \texttt{github.com/VishalPallagani/PLANTS-benchmark}
    };

    \node[fill=boxcolor, text=linkcolor, align=left, rounded corners, draw=none, inner sep=8pt, font=\small, text width=6cm] at (12, 7.5) {
        \faCode \ \href{https://github.com/VishalPallagani/PLANTS-benchmark/blob/main/croissant_metadata.json}{\textbf{Croissant Metadata}} \\
        \texttt{Available on Github (linked above)}
    };

    \node[fill=boxcolor, text=linkcolor, align=left, rounded corners, draw=none, inner sep=8pt, font=\small, text width=6cm] at (4, 3.5) {
        \faFilePdfO \ \href{https://github.com/VishalPallagani/PLANTS-benchmark/blob/main/datasheet-for-PLANTS.md}{\textbf{Datasheet for Dataset}} \\
        \texttt{Available on Github (linked above)}
    };

    \node[fill=boxcolor, text=linkcolor, align=left, rounded corners, draw=none, inner sep=8pt, font=\small, text width=6cm] at (4, 5.5) {
        \faDatabase \ \href{https://huggingface.co/datasets/vishalp/PLANTS-Benchmark}{\textbf{View Data on Huggingface}} \\
        \texttt{huggingface.co/datasets/vishalp/PLANTS-Benchmark}
    };

    \node[fill=boxcolor, text=linkcolor, align=left, rounded corners, draw=none, inner sep=8pt, font=\small, text width=6cm] at (12, 5.5) {
        \faBook \ {\textbf{DOI for Dataset}} \\
        \texttt{doi.org/10.5281/zenodo.11473815}
    };

    \node[fill=boxcolor, text=linkcolor, align=left, rounded corners, draw=none, inner sep=8pt, font=\small, text width=6cm] at (12, 3.5) {
        \faCreativeCommons \ \textbf{License} \\
        \texttt{CC BY 4.0}
    };

\end{tikzpicture}
\end{center}

\section{Dataset Documentation}
We have used the datasheet for datasets \citep{gebru2021datasheets} framework and obtained the associated questionnaire (v7) to document our dataset. We answered the questions in the third person perspective to maintain a formal and objective tone.

\subsection{Motivation}
\noindent \textbf{For what purpose was the dataset created?} 

The PLANTS dataset was created for the novel task of plan summarization, which aims to generate concise summaries of action sequences that achieve specific goals, thereby facilitating quick understanding and decision-making. The authors identified a gap in the existing summarization literature, which has primarily focused on static documents and overlooked dynamic tasks involving sequences of actions, such as workflows, recipes, dialogs, and travel plans. The dataset was designed to address this gap and reinvigorate research in summarization, which some consider a solved problem.

\noindent \textbf{Who created the dataset and on behalf of which entity?}

The dataset was created by Vishal Pallagani and Biplav Srivastava, who are affiliated with the AI Institute at the University of South Carolina.

\noindent \textbf{Who funded the creation of the dataset?} 
There is no funding source for the creation of the PLANTS dataset.

\subsection{Composition}

\noindent \textbf{What do the instances that comprise the dataset represent?}

The instances represent multiple plans to achieve a goal across three categories of planning-like tasks - automated plans (e.g. for intelligent agents), recipes (e.g. cooking cheese sandwich), and travel routes (e.g. driving from Manhattan to Pleasantville).

\noindent \textbf{How many instances are there in total (of each type, if appropriate)?}
There are a total of 30 instances, 10 per each category of the planning-like tasks (automated plans, recipes, and travel routes). Each instance from the automated plans and recipes category has 5 plans, whereas travel routes have 3 plans. Thus, a total of 130 distinct plans are captured in the dataset.

\noindent \textbf{Does the dataset contain all possible instances or is it a sample (not necessarily random) of instances from a larger set?}

Yes, the dataset contains all instances used for benchmark evaluation. The authors also release the generators for the planning-like tasks to create a bigger dataset in the future.

\noindent \textbf{What data does each instance consist of?} 

Each instance has a category, domain, goal, and plans. The category represents the type of planning-like task, the domain (applicable only for automated plans) represents the planning environment with constaints, goal represents the aim of the task, and plan captures different possible plans to achieve the goal. All the instances are represented in text. 

\noindent \textbf{Is there a label or target associated with each instance?}

No, there is no label or target associated with each instance.

\noindent \textbf{Is any information missing from individual instances?}

The dataset does not contain gold plan summaries for each instance as it is challenging to obtain authoritative summaries for planning-like tasks due to their inherent variability and subjective nature.

\noindent \textbf{Are relationships between individual instances made explicit?}

Yes, the plans are for achieving the goal of a particular planning-like task.

\noindent \textbf{Are there recommended data splits (e.g., training, development/validation, testing)?}

Not applicable.

\noindent \textbf{Are there any errors, sources of noise, or redundancies in the dataset?}

No, the dataset being small has been manually verified to check for errors if any.

\noindent \textbf{Is the dataset self-contained, or does it link to or otherwise rely on external resources?}

There is a dependeny on Recipe1M+ dataset to obtain the ingredients and step-by-step instructions for select recipes.

\noindent \textbf{Does the dataset contain data that might be considered confidential?}

No.

\noindent \textbf{Does the dataset contain data that, if viewed directly, might be offensive, insulting, threatening, or might otherwise cause anxiety?}

No.

\noindent \textbf{Does the dataset relate to people?} 

No.

\subsection{Collection process}
\noindent \textbf{How was the data associated with each instance acquired?}

The data is directly observable (raw text).

\noindent \textbf{What mechanisms or procedures were used to collect the data (e.g., hardware apparatus or sensor, manual human curation, software program, software API)?}

For automated plans, SymK planner was used to generate 5 plans per planning problem written in PDDL. For recipes, the authors manually extracted multiple plans to cook a dish from Recipe1M+ dataset. For travel routes, OpenStreetMap API was used.

\noindent \textbf{If the dataset is a sample from a larger set, what was the sampling strategy (e.g., deterministic, probabilistic with specific sampling probabilities)?}

The dataset is not a sample from a larger set.

\noindent \textbf{Who was involved in the data collection process (e.g., students, crowdworkers, contractors) and how were they compensated (e.g., how much were crowdworkers paid)?}

The data collection process is automated.

\noindent \textbf{Over what timeframe was the data collected?}

The dataset was collected from May 10, 2024 to May 15, 2024.

\noindent \textbf{Were any ethical review processes conducted (e.g., by an institutional review board)?}

No, this dataset did not have any sensitive information or human subjects for curation.

\noindent \textbf{Does the dataset relate to people?}

No.

\subsection{Preprocessing/cleaning/labeling}
\noindent \textbf{Was any preprocessing/cleaning/labeling of the data done (e.g., discretization or bucketing, tokenization, part-of-speech tagging, SIFT feature extraction, removal of instances, processing of missing values)?}

Multiple plans that achieve a common goal are represented as a list of lists. No additional preprocessing, cleaning, or labeling of the text data was performed as it was not necessary for our purposes.

\noindent \textbf{Was the “raw” data saved in addition to the preprocessed/cleaned/labeled data (e.g., to support unanticipated future uses)?}

The raw dataset is present in the dataset.

\noindent \textbf{Is the software used to preprocess/clean/label the instances available?}

No preprocessing/cleaning/labeling has been used.

\subsection{Uses}

\noindent \textbf{Has the dataset been used for any tasks already?}

Yes, the dataset has been used for planning task summarization, which involves generating a concise summary of multiple plans that achieve the same goal.

\noindent \textbf{Is there a repository that links to any or all papers or systems that use the dataset?}

\begin{markdown}
Yes, the authors have created a [GitHub](https://github.com/VishalPallagani/PLANTS-benchmark) repository to actively release information related to the task of plan summarization.
\end{markdown}

\noindent \textbf{What (other) tasks could the dataset be used for?}

The dataset is only applicable for the task of plan summarization.

\noindent \textbf{Is there anything about the composition of the dataset or the way it was collected and preprocessed/cleaned/labeled that might impact future uses?}

No.

\noindent \textbf{Are there tasks for which the dataset should not be used?}

The dataset is not fit for any tasks other than plan summarization.

\subsection{Distribution}

\noindent \textbf{Will the dataset be distributed to third parties outside of the entity (e.g., company, institution, organization) on behalf of which the dataset was created?} 

Yes, the dataset is open-sourced and available for research purposes. It is distributed under a Creative Commons (CC) by 4.0 license, allowing third parties to access and use it in accordance with the license terms.

\noindent \textbf{How will the dataset will be distributed (e.g., tarball on website, API, GitHub)?}

\begin{markdown}
The dataset is distributed on GitHub and is available for research purposes. It is also archived and citable with a DOI through [Zenodo](https://doi.org/10.5281/zenodo.11473815).
\end{markdown}

\noindent \textbf{When will the dataset be distributed?}

The dataset is open-sourced on June 04, 2024.

\noindent \textbf{Will the dataset be distributed under a copyright or other intellectual property (IP) license, and/or under applicable terms of use (ToU)?}

The dataset is open-sourced and distributed under a CC-BY-4.0 license.

\noindent \textbf{Have any third parties imposed IP-based or other restrictions on the data associated with the instances?}

No.

\noindent \textbf{Do any export controls or other regulatory restrictions apply to the dataset or to individual instances?}

No.

\subsection{Maintenance}

\noindent \textbf{Who is supporting/hosting/maintaining the dataset?}

The dataset is hosted for free on GitHub. It is maintained by the authors, who ensure that it is kept up-to-date and accessible for research purposes.

\noindent \textbf{How can the owner/curator/manager of the dataset be contacted (e.g., email address)?}

Any details or queries regarding the dataset can be addressed to vishalp@mailbox.sc.edu.

\noindent \textbf{Is there an erratum?}

There are no known errors or corrections related to the dataset at this time. If any errata are identified in the future, they will be documented and accessible through the GitHub repository.

\noindent \textbf{Will the dataset be updated (e.g., to correct labeling errors, add new instances, delete instances)?}

We plan to release a newer version of the dataset in the future with more instances and gold summaries.

\noindent \textbf{If the dataset relates to people, are there applicable limits on the retention of the data associated with the instances (e.g., were individuals in question told that their data would be retained for a fixed period of time and then deleted)?}

No, the dataset does not relate to people.

\noindent \textbf{Will older versions of the dataset continue to be supported/hosted/maintained?}

Yes, all versions of the dataset will be archived on Github as well as Zenodo.

\noindent \textbf{If others want to extend/augment/build on/contribute to the dataset, is there a mechanism for them to do so?}

Yes, the dataset is designed to be extensible. Researchers and developers can use the provided generators to create additional instances or add new planning-like tasks. Contributions can be made by forking the GitHub repository, making the desired changes, and submitting a pull request. We encourage the community to collaborate and enhance the dataset, ensuring it remains a valuable resource for ongoing research.

\section{Plan Summarization Methods}
Here, we provide more details on the three summarization methods - baseline, TextRank, and GPT-4o. 

\subsection{Baseline}
The code for baseline approach is available in the Github repository. The constrains specified in the baseline code are mentioned in Table \ref{tab:constraints}

\begin{table}[h]
\centering
\caption{Constraints and Their Corresponding Values in baseline approach}
\begin{tabular}{>{\raggedright}p{0.4\linewidth} >{\raggedright\arraybackslash}p{0.5\linewidth}}
\toprule
\textbf{Constraint} & \textbf{Value} \\
\midrule
\texttt{ngram\_size} & 3 \\
\texttt{most\_common\_items} & Top 2 unique items \\
\texttt{most\_common\_secondary} & Top 2 secondary mentions \\
\texttt{most\_common\_action\_sequence} & Top 1 most common 3-gram sequence \\
\texttt{most\_common\_roads} & Top 5 most common roads \\
\bottomrule
\end{tabular}
\label{tab:constraints}
\end{table}

\subsection{TextRank}
The code for TextRank is also made available in the Github repository. The constraints followed for TextRank are similar to Table \ref{tab:constraints}. 

\subsection{GPT-4o}
The prompt template used for GPT-4o in the domain of automated plans, along with the responses, can be accessed through this \href{https://chatgpt.com/share/22feef7f-3b0c-4070-8076-f78462518c60}{\textbf{link}}. A similar prompt template is used for both recipes, as well as travel routes.

For GPT-4o, we could not find a prompt that made the model adhere to the constraints. In our experimentation, we found GPT-4o's outputs to be very verbose, regardless of the prompt instructions for the constraints. In our experiments, we use the following textual description as part of the prompt for defining the constraints:

\begin{tcolorbox}[colback=blue!5!white, colframe=blue!75!black, width=\textwidth, boxrule=0.5mm, sharp corners]
\texttt{The summary should be constrained by the text's length and the number of actions (make sure to capture captions of high importance for a plan).}
\end{tcolorbox}

\section{Plan Summarization Outputs}

We consider the PL task of recipes with the goal of making a cheese sandwich. There are five different recipes in the PLANTS dataset to prepare a cheese sandwich, shown in Table \ref{tab:cheese_sandwich_recipes} which are provided as inputs to the three summarization methods. The output summaries obtained for cheese sandwich recipes from three different methods are shown in Table \ref{tab:summaries}. The outputs from the three different methods for all the instances in PLANTS is in the Github repository.

\begin{longtable}{>{\raggedright}p{0.05\linewidth}|>{\raggedright}p{0.45\linewidth}|>{\raggedright\arraybackslash}p{0.45\linewidth}}
\caption{Different recipes for preparing a cheese sandwich}\label{tab:cheese_sandwich_recipes} \\
\toprule
\textbf{ID} & \textbf{Ingredients} & \textbf{Instructions} \\
\midrule
\endfirsthead
\multicolumn{3}{c}{\textbf{Table continued from previous page}} \\
\toprule
\textbf{ID} & \textbf{Ingredients} & \textbf{Instructions} \\
\midrule
\endhead
\midrule
\multicolumn{3}{r}{\textbf{Table continued on next page}} \\
\endfoot
\bottomrule
\endlastfoot

1 & 
2 slices sourdough bread, 2 tbsp salted butter, ⅓ cup vintage cheddar cheese, ⅓ cup mozzarella. & 
1. Butter bread – Slather both sides of each piece of bread with the butter (yes, all 4 sides!). \newline
2. Light toast – Heat a heavy-based skillet or frying pan over medium low heat (no oil or butter). Place both pieces of bread in the skillet and lightly toast for 1 minute to warm it through and create a light crust. (When we flip, this gives the cheese a head start). \newline
3. Pile on cheese – Flip one slice of bread, then pile on the cheddar cheese followed by the mozzarella. Place the other slice of bread on top, with the hot toasted side in contact with the cheese. \newline
4. Cook 3 minutes – Cook for 3 minutes or until the bread is evenly golden and crisp, pressing down lightly with a spatula every now and then. If it's browning too quickly, remove from the stove to cool down a bit and lower heat. \newline
5. Flip, 3 minutes – Turn the sandwich over, and cook the other side for 3 minutes or until the bread is golden and the cheese is melted. \newline
6. Devour – Transfer to cutting board. Cut in half in one swift motion. Admire cheese pull. Devour. \\
\midrule

2 & 
4 slices bread of your choice, 4 slices mozzarella cheese, 2 tablespoons unsalted butter, ½ teaspoon oregano, ½ to 1 teaspoon chili flakes, ½ teaspoon black pepper. & 
1. Butter each bread slice on both the sides. Heat a pan and place 2 buttered bread slices on it. \newline
2. Grill until golden. Flip the slice and place/spread the cheese on the hot bread. Sprinkle chili flakes and herbs, if you want. \newline
3. Cover the bread with another slice with the hotter side over the cheese. \newline
4. On a low to medium heat, grill it until golden and the cheese begin to melt. Halve the sandwiches and serve. \\
\midrule

3 & 
4 slices bread of your choice, 4 slices mozzarella cheese, 2 tablespoons unsalted butter, ½ teaspoon oregano, ½ to 1 teaspoon chili flakes, ½ teaspoon black pepper. & 
1. Start with your oven set to 450 F or 220 C. Or if using a sandwich toaster, preheat it. \newline
2. Generously butter your bread and assemble your sandwiches, by placing the buttered side down on a baking sheet or a tray. \newline
3. Place the cheese slices and add your spices or herbs of choice, if you want to. Cover with another bread slice, with the buttered side on top. \newline
4. Place tray on the middle rack of your oven and toast until the bread begins to turn golden brown, about 3 to 5 mins. If you are using a OTG, it just takes about 3 to 4 mins. \newline
5. Flip once with a wooden spatula and continue to grill until the cheese is visibly melted and your sandwich has developed an even, light brown color. This takes another 3 to 5 mins depending on your bread and oven. \newline
6. Remove and let cool down on a wired rack before you cut and enjoy hot! \\
\midrule

4 & 
4 slices bread of your choice, 4 slices mozzarella cheese, 2 tablespoons unsalted butter, ½ teaspoon oregano, ½ to 1 teaspoon chili flakes, ½ teaspoon black pepper. & 
1. Place the assembled sandwiches in the air fryer and air fry at 350F or 180 C for 5 mins. Do not air fry for too long, the edges turn hard and too crisp. \\
\midrule

5 & 
18 slices bread, 4 tablespoons butter, 9 slices cheddar cheese. & 
1. Preheat the oven to 450 degrees F (230 degrees C). \newline
2. Butter one side of 9 slices of bread, and place butter-side down on a baking sheet. Arrange cheese on each slice of bread. Spread butter on 9 remaining slices of bread, and place them buttered-side up on top of the cheese. \newline
3. Bake in the preheated oven for 6 to 8 minutes. Flip the sandwiches, and bake an additional 6 to 8 minutes, or until golden brown. \\
\end{longtable}

\begin{longtable}{|p{0.25\linewidth}|p{0.8\linewidth}|}
\caption{Summaries of grilled cheese sandwich recipes from different methods} \label{tab:summaries} \\
\toprule
\textbf{Method} & \textbf{Summary} \\
\midrule
\endfirsthead
\multicolumn{2}{c}{\textbf{Table continued from previous page}} \\
\toprule
\textbf{Method} & \textbf{Summary} \\
\midrule
\endhead
\midrule
\multicolumn{2}{r}{\textbf{Table continued on next page}} \\
\endfoot
\bottomrule
\endlastfoot

\textbf{Baseline} & 
Analysis of 5 recipes reveals an average of 5.00 steps per recipe, with must-have ingredients - 4 slices mozzarella cheese (3 times), 2 tablespoons unsalted butter (3 times), and key actions - ('butter slice', 1) (1 times), ('heat pan', 1) (1 times). The recipe with the least number of steps is Recipe: 4. \\
\midrule

\textbf{TextRank} & 
Must-have Ingredients: \newline
4 slices bread of your choice (3 times) \newline
4 slices mozzarella cheese (3 times) \newline
2 tablespoons unsalted butter (3 times) \newline
½ teaspoon oregano (3 times) \newline
½ to 1 teaspoon chili flakes (3 times) \newline
\newline
Key Instructions: \newline
3. Place the cheese slices and add your spices or herbs of choice, if you want to. Cover with another bread slice, with the buttered side on top. \newline
1. Butter each bread slice on both the sides. Heat a pan and place 2 buttered bread slices on it. \newline
2. Grill until golden. Flip the slice and place /spread the cheese on the hot bread. Sprinkle chili flakes and herbs, if you want. \newline
5. Flip, 3 minutes – Turn the sandwich over, and cook the other side for 3 minutes or until the bread is golden and the cheese is melted. \newline
2. Butter one side of 9 slices of bread, and place butter-side down on a baking sheet. Arrange cheese on each slice of bread. Spread butter on 9 remaining slices of bread, and place them buttered-side up on top of the cheese. \newline
3. Pile on cheese – Flip one slice of bread, then pile on the cheddar cheese followed by the mozzarella. Place the other slice of bread on top, with the hot toasted side in contact with the cheese. \\
\midrule

\textbf{GPT-4o} & 
Summary of Grilled Cheese Sandwich Recipes: \newline
\textbf{Ingredients:} \newline
Bread: Sourdough, or bread of choice (2-4 slices per sandwich) \newline
Cheese: Cheddar, mozzarella, or a combination (2-4 slices or 1/3 cup shredded per sandwich) \newline
Butter: Salted or unsalted, 2 tablespoons to spread on both sides of the bread \newline
Optional Seasonings: Oregano, chili flakes, black pepper \newline
\newline
\textbf{Instructions:} \newline
1. Prepare Bread and Cheese: Butter both sides of each bread slice. For extra flavor, sprinkle seasonings like oregano, chili flakes, and black pepper on the cheese. \newline
2. Assemble Sandwich: Place one slice of bread buttered-side down in a skillet, on a baking sheet, or in an air fryer. Add cheese on top, then cover with another slice of bread, buttered-side up. \newline
3. Cook: \newline
\textbf{Stovetop:} Heat a skillet over medium heat. Place the sandwich in the skillet and cook for 3 minutes on each side, pressing down lightly with a spatula until golden brown and cheese is melted. \newline
\textbf{Oven:} Preheat to 450°F (230°C). Place the sandwich on a baking sheet and bake for 6-8 minutes per side, until golden brown. \newline
\textbf{Air Fryer:} Preheat to 350°F (180°C). Air fry the sandwich for 5 minutes, ensuring it doesn't get too crisp. \newline
4. Serve: Cut the sandwich in half and serve hot, enjoying the melted cheese and crispy bread. \newline
This summary captures the essential steps and common ingredients across the provided grilled cheese sandwich recipes while ensuring conciseness and clarity. \\
\end{longtable}

\section{User Study Visualizations}

\subsection{Ease of understanding of the summaries}
\begin{figure}[h]
    \centering
    \subfigure[Results for automated plans]{
        \includegraphics[width=0.7\textwidth]{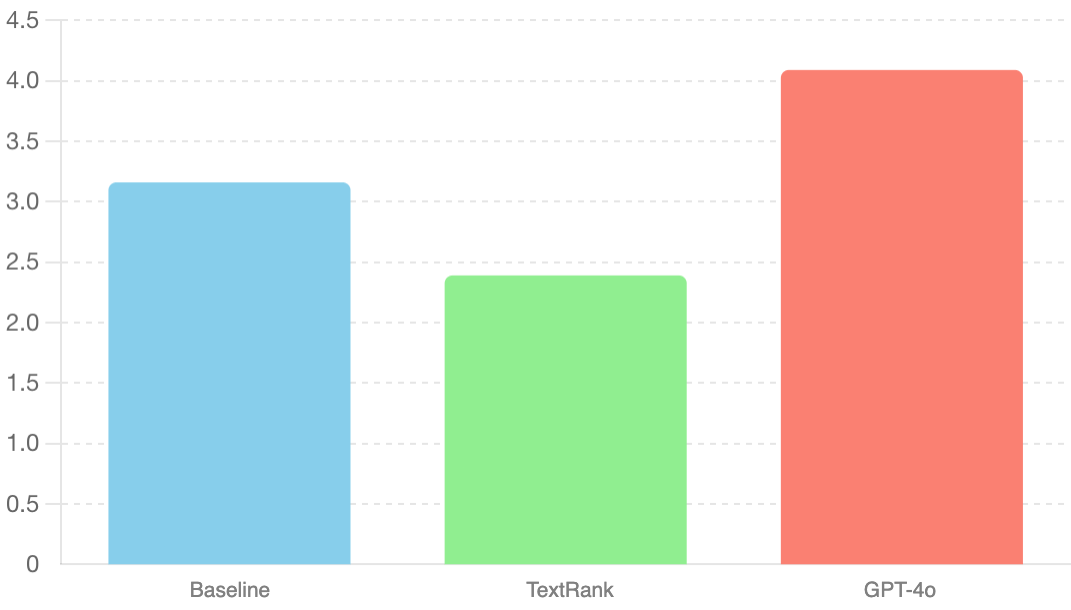}
        \label{fig:subfig1}
    }
    \subfigure[Results for recipes]{
        \includegraphics[width=0.7\textwidth]{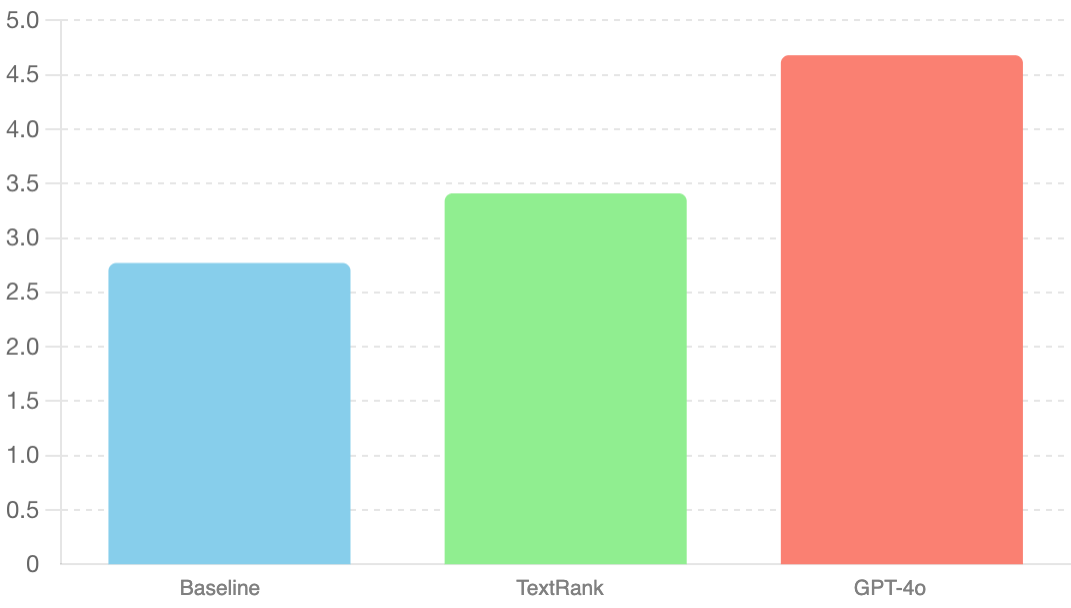}
        \label{fig:subfig2}
    }
    \subfigure[Results for travel routes]{
        \includegraphics[width=0.7\textwidth]{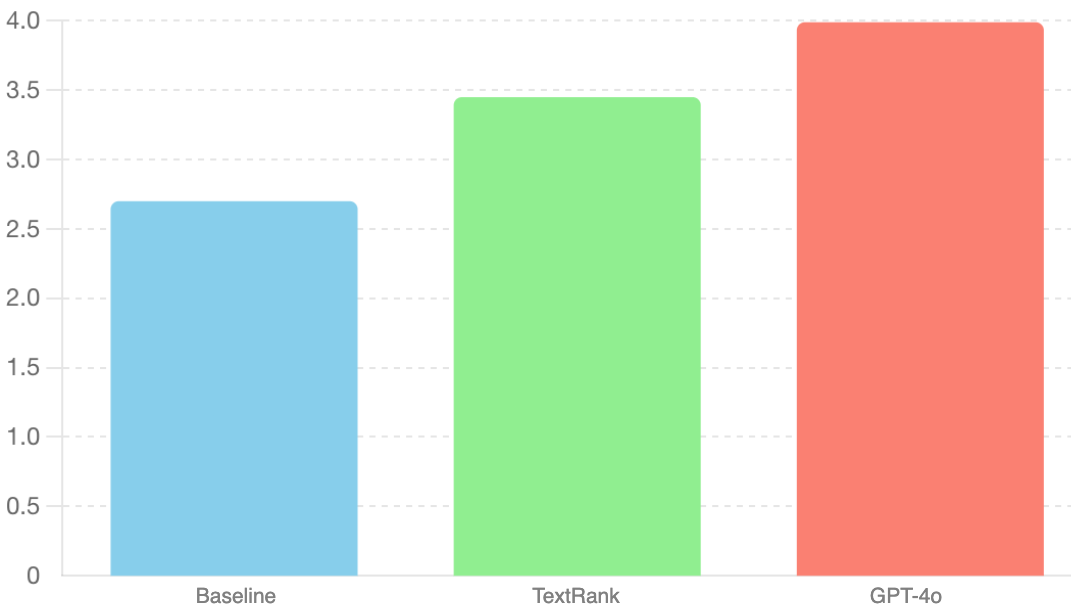}
        \label{fig:subfig3}
    }
    \caption{GPT-4o is chosen as the easy to understand summary for all PL tasks, while the baseline approach and TextRank alternate between second and third positions.}
    \label{fig:subfigures}
\end{figure}

\bibliographystyle{plainnat}
\bibliography{references.bib}